# Expanding Vietnamese SentiWordNet to Improve Performance of Vietnamese Sentiment Analysis Models

Hong-Viet Tran [1]    Van-Tan Bui [2*]    Lam-Quan Tran[3]

[1]*University of Engineering and Technology, Vietnam National University, Hanoi, Vietnam*
[2]*University of Economic and Technical Industries, Hanoi, Vietnam*
[3]*Hanoi University of Public Health, Hanoi, Vietnam*
\* Corresponding author's Email: bvtan@uneti.edu.vn

**Abstract:** Sentiment analysis is one of the most crucial tasks in Natural Language Processing (NLP), involving the training of machine learning models to classify text based on the polarity of opinions. Pre-trained Language Models (PLMs) can be applied to downstream tasks through fine-tuning, eliminating the need to train the model from scratch. Specifically, PLMs have been employed for Sentiment Analysis, a process that involves detecting, analyzing, and extracting the polarity of text sentiments. Numerous models have been proposed to address this task, with pre-trained PhoBERT-V2 models standing out as the state-of-the-art language models for Vietnamese. The PhoBERT-V2 pre-training approach is based on RoBERTa, optimizing the BERT pre-training method for more robust performance. In this paper, we introduce a novel approach that combines PhoBERT-V2 and SentiWordnet for Sentiment Analysis of Vietnamese reviews. Our proposed model utilizes PhoBERT-V2 for Vietnamese, offering a robust optimization for the prominent BERT model in the context of Vietnamese language, and leverages SentiWordNet, a lexical resource explicitly designed to support sentiment classification applications. Experimental results on the VLSP 2016 and AIVIVN 2019 datasets demonstrate that our sentiment analysis system has achieved excellent performance in comparison to other models.

## 1. Introduction

Social space on the Internet brings many positive values for the development of society and individuals. The amount of information data also becomes huge and constantly increasing. Efficient exploitation of that amount of data is a core problem of AI, which brings superior performance to AI systems. Along with the development of the internet, especially the evelopment of social networking sites (Facebook, Instagram, Twitter ...) and e-commerce sites (Lazada, Shopee, Tiki ... the amount of data transmitted through it has become enormous and constantly increasing. This development has allowed people not only to share information but also to express their attitudes and opinions towards products, services, and other social issues. Therefore, we focus on research sentiment analysis to improve the quality of products and services.

In this study, we introduce a new approach to build a model to Vietnamese reviews for Sentiment Analysis, which is based on combining PhoBERT [1] and SentiWordnet [2]. We study a sentiment analysis model using PhoBERT pre-trained model for Vietnamese, and SentiWordNet, a lexical resource explicitly devised for supporting sentiment classification and opinion mining applications. The rest of this paper is organized as follows: Related works for the sentiment analysis task are mentioned in Section 2. Section 3 gives an overview of PhoBERT model. Section 4 describes details of our methods for combining PhoBERT and SentiWordNet on sentiment analysis. The experimental results and analysis with our classification approach are described in Section 5. Finally, Section 6 concludes the paper.

## 2. Related Work

Sentiment analysis systems were used for analyzing customer reviews, which aim to understand the public's point of view. These opinions may be consisting of positive or negative reviews toward specific products [3]. According to the traditional approach, several social media platforms were applied to analyze products and determine customer opinion [4]. Besides, there have also been sentiment analysis studies focused on understanding and identifying linguistic markers in extremist web forum discussion [5]. This extremism refers to opinions or statements that mean to perpetuate hate or incite violence. Extremist forums are commonly studied targets as they not only can to examine extremist discourse and attitudes [6] but also can contain terrorists who can be identified using sentiment analysis [7].

Sentiment Analysis is the process of analyzing and evaluating a person's point of view on an object (negative, positive, or normal opinion, …). This process can be done by using rule-based sets, using machine learning, and lexicon/dictionary approaches. [8] research to classify the reviews into two groups, positive and negative. Many classifying techniques in machine learning as Support vector machines(SVM) [9], Native Bayes [10], ... are used to solve this problem. [11] uses a dictionary of emotional vocabulary to indicates whether a word is positive or negative along with its level.

For text classification tasks like sentiment analysis, deep learning methods enhance word representations by capturing contextual meaning. Mikolov and colleagues introduced an unsupervised neural network algorithm, "Paragraph vector [12]. Zhang and colleagues utilized Convolutional Neural Networks (CNNs) for document classification, operating at the character level [13]. These approaches use matrices of text for input, where each column represents a "Paragraph vector", aiming to extract sentiment from semantic cues in comments or documents.

Recent studies have integrated deep learning (DL) with traditional methods to boost sentiment analysis performance [14]. CNNs excel in word-level tasks without needing explicit syntactic or semantic knowledge [15], while RNNs effectively handle sequential data [16]. A three-way CNN model (3W-CNN) combines DL and traditional ML algorithms[17], demonstrating improved accuracy. DL architectures, including CNN and RNN along with its variant LSTM, outperform others in sentiment analysis tasks across multiple languages[18].

Researchers have explored various DL architectures like CNNs and RNNs for sentiment analysis [19], [20]. Using Word2vec embeddings, CNN models can effectively analyze public opinions with enhanced accuracy [21]. Muatheer framework applied SA techniques to identify influential social media users, achieving significant success [22].

## 3. Background

### 3.1. PhoBERT pre-trained model

BERT (Bidirectional Encoder Representations from Transformers) is a multi-layered structure of Bidirectional Transformer encoder layers, based on the architecture of transformer [23]. This language representation model has successfully improved in finding representations of words in digital space from their context. BERT uses Bidirectional Transformer encoders to replace Encoders combining Decoders and BERT fine-tuning tasks do not require Decoder blocks. BERT is to apply two-dimensional training techniques of Transformers from a very famous Attention model to a Language Model.

PhoBERT[1] [1], Pre-trained language models for Vietnamese are the state-of-the-art language models for Vietnamese. Two PhoBERT versions of *base* and *large* are the first public large-scale monolingual language models pre-trained for Vietnamese. PhoBERT pre-training approach is based on RoBERTa [24] which optimizes the BERT pre-training procedure for more robust performance. PhoBERT outperforms previous monolingual and multilingual approaches, obtaining new state-of-the-art performances on four downstream Vietnamese NLP tasks of Part-of-speech tagging, Dependency parsing, Named-entity recognition, and Natural language inference.

There are two approaches for the sentiment analysis task of review texts in the followings:

Feature extraction: in this approach, PhoBERT is used as a feature extraction model. The architecture of the PhoBERT model is preserved. PhoBERT model's outputs are feature input vectors for subsequent classification models to solve the given problem.

Fine-tuning: the architecture of the model is modified by adding some layers at the end of PhoBERT model. These layers will solve the problem, and then retrain the model. This method is used to evaluate benchmarks on different tasks, showing PhoBERT superiority over previous models.

### 3.2. Long Short-Term Memory Architecture

The recurrent neural network (RNN) is suitable for modeling sequential data by nature, as it keeps a hidden state vector h, which changes with input data at each step accordingly. One problem of RNN model is known as gradient vanishing or exploding. Long short term memory (LSTM) units are proposed by Hochreiter and Schmidhuber [25] to overcome this problem. The main idea is to introduce an adaptive gating mechanism, which decides the degree to which LSTM units keep the previous state and memorize the extracted features of the current data input. LSTM models seem well-suited for modeling sequential data by a vector representation. The LSTM-based recurrent neural network comprises four components: an input gate $i_t$, a forget gate $f_t$, an output gate $o_t$, and a memory cell $c_t$. The three adaptive gates $i_t$, $f_t$ and $o_t$ depend on the previous state $h_{(t-1)}$ and the current input $x_t$ (Equations 1). An extracted feature vector $g_t$ is also computed, by Equation 2, serving as the candidate memory cell.

---
[1] https://github.com/VinAIResearch/PhoBERT

$$i_t = \sigma(W_i \cdot x_t + U_i \cdot h_{t-1} + b_i)$$
$$f_t = \sigma(W_f \cdot x_t + U_f \cdot h_{t-1} + b_f) \quad (1)$$
$$o_t = \sigma(W_o \cdot x_t + U_o \cdot h_{t-1} + b_o)$$
$$g_t = tanh(W_i \cdot x_t + U_i \cdot h_{t-1} + b_i) \quad (2)$$

The current memory cell $c_t$ is a combination of the previous cell content $c_{t-1}$ and the candidate content $g_t$, weighted by the input gate $i_t$ and forget gate $f_t$, respectively (Equation 3).

$$c_t = i_t \otimes g_t + f_t \times c_{t-1} \quad (3)$$

The output of LSTM units is the the recurrent network's hidden state, which is computed by Equation 4 as follows.

$$h_t = o_t \otimes tanh(c_t) \quad (4)$$

In the above equations, $\sigma$ denotes a sigmoid function; $\otimes$ denotes element-wise multiplication.

### 3.3. Recurrent Convolutional Neural Networks

The Recurrent Convolutional Neural Network architecture (RCNN) was proposed by Siwei Lai and colleagues for text classification tasks [26]. The model utilizes a hybrid architecture integrating recurrent and convolutional components. Specifically, it employs two LSTM networks: one for capturing sequential context from left to right, and another for contextual understanding from right to left. Subsequently, the outputs from both LSTM networks are further processed through a conv1d layer. This convolutional layer plays a crucial role in extracting higher-level features from the sequential representations provided by the LSTM networks, thereby enhancing the model's capability to learn complex patterns in the data.

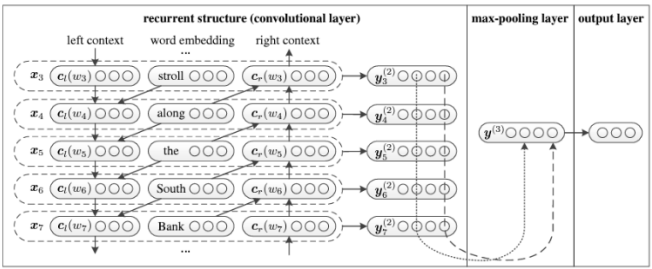

*Figure 1: The structure of the recurrent* convolutional

*neural network* [26].

## 4. Methodology

In this paper, we propose a new approach to combine PhoBERT and SentiWordnet for Sentiment Analysis of Vietnamese reviews. We study a sentiment analysis model using PhoBERT pre-trained model for Vietnamese, which is a robust optimization for Vietnamese of the prominent BERT model, and SentiWordNet, in which a lexical resource explicitly devised for supporting sentiment classification applications. The **Error! Reference source not found.** depicts the architecture of our proposed model.

### 4.1. SentiWordNet

SentiWordNet is a natural language resource developed based on WordNet [27], a lexical database of synonymous and related words, combined with information about the emotional or sentiment-related meanings of the words. SentiWordNet [2] assigns a triad of numerical values to each word in WordNet, often represented as ratios, indicating the degree of various emotions. Specifically, these three values are typically positive, negative, and neutral. It was made publicly available for research purposes.

SentiWordNet is pivotal for sentiment analysis tasks due to its fine-grained sentiment scoring for individual words. This granularity allows for nuanced lexical analysis, capturing subtle emotional nuances. Additionally, its provision of sentiment scores for each word sense aids in disambiguating polysemous terms, enhancing sentiment classification accuracy across different contexts. Moreover, SentiWordNet facilitates domain adaptation by supplementing domain-specific sentiment lexicons and seamlessly integrating with existing NLP frameworks. Its structured representation of sentiment information empowers sentiment analysis systems to achieve heightened precision and efficiency in analyzing textual data.

A SentiWordNet dictionary is constructed through two stages as follows: initially, WordNet lexical-semantic relations such as synonym, antonym, and hyponymy are explored to extend a core of seed words used in [28], and known a priori to carry positive or negative opinion bias. After a fixed number of iterations, a subset of WordNet words is obtained with either a positive or negative label. These word glosses are then used to train a committee of machine learning classifiers. The classifiers are trained with different algorithms and different training set sizes to minimize bias. predictions from the classifier committee are used to determine the sentiment orientation of the remainder of terms in WordNet.

To improve the performance of Vietnamese sentiment analysis models, Xuan-Son Vu et al proposed Vietnamese SentiWordNet [29]. This dictionary includes 1,017 Vietnamese words labeled with part-of-speech and word definitions (SynsetTerms Gloss). Each word is evaluated positive score (PosScore) and negative score (NegScore).

A. Expanding Vietnamese SentiWordNet

For the Sentiment Analysis task in Vietnamese, Xuan-Son Vu and colleagues proposed a method to construct a SentiWordNet dictionary for Vietnamese [29]. The Vietnamese SentiWordNet was subsequently employed and demonstrated to be a valuable lexical resource for enhancing the performance of Vietnamese Sentiment Analysis models. The Vietnamese SentiWordNet (ViSentiWordNet) currently comprises only 1,017 words, including nouns, adjectives, and verbs. A limitation of ViSentiWordNet is its small size, which results in insufficient coverage of the vocabulary. In this study, we propose a method to extend ViSentiWordNet to increase its coverage of the Vietnamese lexicon (ExtViSentiWordNet). To expand the ViSentiWordNet dictionary, we undertake the following steps:

Step 1: We extract from SentiWordNet the set of positive words $P$. The set $P$ includes words with a $positive\ score > T$ and a $negative\ score = 0$. Similarly, the set of purely negative words $N$ includes words with a $negative\ score > T$ and a $positive\ score = 0$. The threshold $T$ is a constant determined through experimentation.

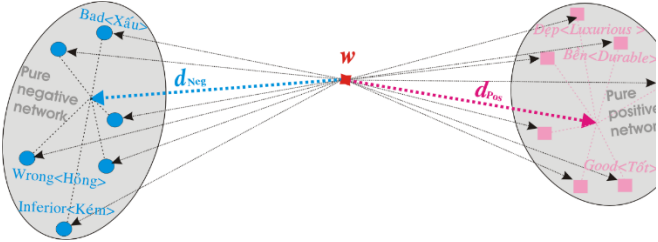

Figure 2: Schema of Vietnamese SentiWordNet extension method.

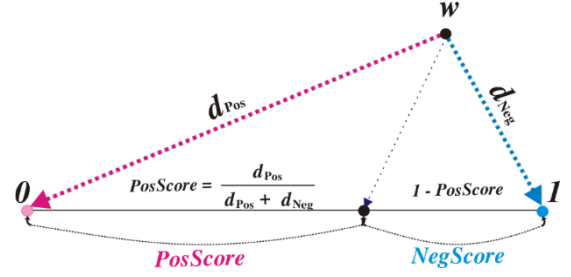

Figure 3: Illustrating the method of computing the positive and negative scores of a word w.

Step 2: We expand the sets $P$ and $N$ according to the following rules: we select synonyms from the VCL and WordNet dictionaries. Synonyms of words in $P$ are added to the set $P$, and synonyms of words in $N$ are added to the set $N$. Similarly, antonyms of words in $P$ are added to the set $N$, and antonyms of words in $N$ are added to the set $P$ (Figure 2).

Step 3: To add the word w to the SentiWordNet dictionary, we calculate the average distance dpos from $w$ to words in set $P$, and dneg as the average distance from $w$ to words in set $N$. The positive score of $w$ is calculated using the Equation 5. Figure 4 illustrates visually the computation of the positive score and negative score for a word $w$.

$$PosScore = \frac{d_{pos}}{d_{pos} + d_{neg}} \quad (5)$$

The negative score is calculated according to the following equation:

$$NegScore = 1 - PosScore \quad (6)$$

**4.2. Exploiting Vietnamese SentiWordNet**

In this study, we exploit the extended Vietnamese Senti-WordNet to extract feature vectors representing the positive and negative aspects of review text segments. Accordingly, we propose the $SentiVector\_Extraction$ algorithm (Algorithm 1: The algorithm utilizes Vietnamese SentiWordNet to extract PosVec and NegVec vectors.) to extract two vectors, PosVec and NegVec, from each review text segment. Here, PosVec encodes positive sentiment information of the review text segment, whereas NegVec en_codes negative sentiment information. Additionally, we define a set of negation words such as *vô<not>, bất<not>, chẳng<not>, không<not>, kém<less>, chẳng hề < not at all>, không bao giờ<never>, chẳng bao giờ<never>* which are used to form negation patterns. Positive words within these patterns often imply negative meanings, whereas negation words in these patterns imply positive meanings. In Algorithm 1, the Vector Size Normalization Operator (VSNO) is employed to normalize the sizes of PosVec and NegVec vectors, and ⊕ denotes the vector concatenation operator.

For example, although *đẹp<beautiful>, tốt<good>, bền<durable>, rẻ<cheap>, tài<talented>* are positive words, but *không đẹp<not beautiful>, chẳng đẹp<not pretty>, không bền<not druable>, không tốt<not good>, không rẻ<not cheap>, bất tài<incompetent>* having negative meanings. Considering two sentences:

*S*1: *Chiếc xe này chạy **nhanh** mà còn **tiết kiệm** nhiên liệu nữa.* <The car is not only fast but also fuel-efficient.>

*S*2: *Chiếc xe này không **nhanh** mà còn chẳng **tiết kiệm** nhiên liệu nữa.*<The car is not fast and also not fuel-efficient.>

It can be seen intuitively that $S1$ is semantically positive but $S2$ tends to be semantically negative.

Algorithm 1: The algorithm utilizes Vietnamese SentiWordNet to extract PosVec and NegVec vectors.

*Algorithm SentiVector_Extraction(S, SeWN)*
*Input: S (data sample), SeWN (Vietnamese SentiWordNet)*
*Output: PosVec, NegVec (sentiment feature vectors)*

*1. Initialize PosVec ← [], NegVec ← []*
*2. For each w in S do*
   *If w exists in SeWN then*
     *If w matches a reverse pattern then*
       *Append SeWN[w].PosScore to NegVec*
       *Append SeWN[w].NegScore to PosVec*
     *Else*
       *Append SeWN[w].PosScore to PosVec*
       *Append SeWN[w].NegScore to NegVec*
     *End If*
   *End If*
  *End For*
*3. Normalize PosVec using DVSNO(PosVec)*
*4. Normalize NegVec using VSNO(NegVec)*
*5. Return PosVec, NegVec*
*End Algorithm*

### 4.3. Combining phoBERT and SentiWordNet Feature

In this study, we propose a combined model termed Neural Network leveraging Language Model and SentiWordNet vectors for Sentiment Analysis (CombViSA). The LSSA model exploits the vector representations of reviews through the phoBERT-V2 model combined with selected vectors from ExtSentiWordNet. For each review, we utilize the complete output of phoBERT-V2, including the token [CLS]. These outputs constitute an $SL \times h$ matrix, where $SL$ represents the maximum length of the input sequence, and $h$ denotes the length of hidden vectors. Subsequently, this output matrix serves as input to the RCNN module. The output vector from the RCNN module is passed through a Feedforward neural network module to obtain the LMVec vector. Furthermore, for each review, we extract two feature vectors based on ExtSentiWordNet, namely PosVec and NegVec. These two vectors are concatenated and fed into a Feedforward neural network module to obtain SWVec. The LMVec vector is then concatenated with the SWVec vector and passed into another Feedforward neural network module. Finally, we employ the Softmax function to obtain the classification results as described in Equation 7.

$$Softmax(MLP(LMVec \oplus SWVec)) \quad (7)$$

In Equation 7, the operator $\oplus$ denotes the concatenation of two vectors.

## 5. Experimental Results

### 5.1. Datasets

In this study, we empirically evaluate the proposed model using two Vietnamese datasets. These datasets consist of reviewing texts commented on by users on e-commerce sites.

• Firstly, the VLSP 2016 [2] Shared Task on Sentiment Analy-sis provides the dataset for this study. The data comprises user reviews of technological devices categorized into three sentiment classes: "negative", "positive", and "neutral". Given that a review may express complex sentiments about various aspects, we have applied the following constraints to the dataset:

- The dataset exclusively comprises reviews that con-tain personal opinions.

- The data typically consist of brief comments that provide opinions about a single object. There are no constraints regarding the number of aspects of the object addressed in the comment.

- The label (positive, negative, or neutral) represents the overall sentiment of the entire review.

The dataset exclusively comprises genuine data col-lected from social media and does not include any data artificially generated by humans.

• Secondly, the AIVIVN 2019 [3] dataset is constructed from the training data used in the AIVIVN sentiment analysis competition. This dataset comprises user comments on product reviews from various e-commerce sites. The data is categorized into two sentiment labels: "positive" and "negative" Employing two datasets with different characteristics would help to better evaluate the performance of different models.

---

[2] https://vlsp.org.vn/resources-vlsp2016

[3] https://www.aivivn.com/contests/1

Table 1. Data description after pre-process.

| Dataset | Train | | Test | | Totally |
|---|---|---|---|---|---|
| | Positive | Negative | Positive | Negative | |
| VLSP 2016 | 20,439 | 20,267 | 5,000 | 5,000 | 50,760 |
| AIVIVN 2019 | 22,979 | 19,537 | 8,301 | 6,795 | 57,612 |

Some data statistics are made for the evaluation process are showed in Tables 1 and Table 2. These are the statistics after the data have been processed. An important data feature of the sentiment analysis problem is that comments are often grammatically and lexicographically irregular, and they contain many special symbols. It will be a challenge for models to understand such messages. Utilizing two datasets with distinct characteristics can enhance the evaluation of various models' performance. Statistics pertinent to the evaluation process are presented in Tables 1 and 2. These statistics reflect the data post-processing. A key aspect of sentiment analysis is that comments frequently exhibit grammatical and lexical irregularities and include numerous special symbols. Such characteristics present a challenge for models in comprehending these messages.

Table 2. Statistics of words included in the comments.

| | VLSP 2016 | AIVIVN 2019 |
|---|---|---|
| Mean | 25.34 | 55.45 |
| Stdn | 29.67 | 63.75 |
| Min | 1 | 1 |
| 25% | 9 | 14 |
| 50% | 16 | 32 |
| 75% | 31 | 76 |
| Max | 631 | 435 |

Table 3. Experimental results on VLSP 2016 datasets.

| Model | Precision | Recall | F1 |
|---|---|---|---|
| SA1 [31] | - | - | 0.80 |
| SA2 [31] | - | - | 0.71 |
| phoBERT-V2 | 0.92 | 0.90 | 0.91 |
| phoBERT-V2+ExtViSentiWordNet | 0.94 | 0.92 | 0.93 |
| **CombViSA** | **0.96** | **0.94** | **0.95** |

### 5.2. Models

In this study, we carry out experiments with the proposed model, which combines PhoBERT and VietSentiWordNet. The performance of the model are evaluated on two datasets VLSP 2016 and AIVIVN 2019, using the measures Precision, Recall, and F1 score (F1). The performance of our model is compared with models including using PhoBERT only, Multilingual BERT-base, as well as some experimental results reported in [30], specifically as follows.

• **FastText + LSTM, Glove + LSTM, SA1**, and **SA2**: To evaluate the performance of the proposed model, we refer to the experimental results of these two models which are reported in [30]. Two models **SA1** and **SA2** are reported in [31].

• **phoBERT-V2:** This model exploits only the phoBERT-base model. The representation vectors of the top four layers are concatenated to push into a multilayer perceptron module to generate the classification result.

• **phoBERT-V2+ExtViSentiWordNet:** This model exploits the phoBERT-V2 model and the Senti vector which is derived from ExtViSentiWordNet. Senti vector is coupled with feature vector obtained from PhoBERT, combined feature vector is pushed into multilayer perceptron mod-ule to get prediction results.

### 5.3. Parameter Settings

In this study, we employ the phoBERT-V2 model to generate vectors representing the words in comments. To standardize the dimensions of the PosVec and NegVec vectors, we set the maximum number of words per comment to 128, resulting in both PosVec and NegVec having 128 dimensions. Several hyperparameters were configured for training the model as follows: the batch size was set to 24, the number of epochs was set to 20, the accumulation steps parameter was set to 16, and the learning rate was set to 3e-5.

### 5.4. Results and Discussion

We evaluate the proposed model on two datasets: VLSP 2016 and AIVIVN 2019. The performance results of all competing models on the VLSP 2016 dataset are presented in Table 3. The experimental results indicate that the phoBERT model achieves significantly higher performance compared to the baseline models, which are the top models reported in [30]. Additionally, integrating SentiVec as lexical-semantic features with phoBERT-V2 features leads to a substantial improvement in model performance.

Table 4. experimental results on aivivn 2019 dataset

| Model | Precision | Recall | F1 |
|---|---|---|---|
| FastText + LSTM [30] | 0.88 | 0.86 | 0.87 |
| Glove + LSTM [30] | 0.85 | 0.85 | 0.85 |
| phoBERT-V2 | 0.88 | 0.90 | 0.89 |
| phoBERT-V2+ExtViSentiVec | **0.93** | **0.92** | **0.93** |
| **CombViSA** | **0.95** | **0.94** | **0.94** |

Experiments with the AIVIVN 2019 dataset also show that the PhoBERT-V2 model has slightly better performance than the two baseline models that used LSMT with GloVe, fastText. The phoBERT-V2 + ViSentiVec model which exploited SentiVec achieved significantly higher performance than the original PhoBERT model as well as the baseline models.

As shown in Table and Table , experimental results showed that our system achieved high performance in comparison to the baseline models, the best result of the proposed model is 0.92 in terms of F1-score. As shown in Table , the average length of comments is quite large, 23.39 for VLSP 2016, and 55.45 for AIVIVN 2019. Since LSTM is inherently inefficient for long sentences, models using LSTM achieve low performance on datasets containing long comments. Experimental results show that: Firstly, ViSentiWordNet is a useful lexical-semantic feature to improve the performance of the sentiment analysis problem. Secondly, the phoBERT-V2 model is more suitable for solving this problem than other word embedding models.

Although SentiVec has proven effective for sentiment analysis tasks, a significant limitation of ViSentiWordNet is its coverage. The ViSentiWordNet contains just over 1,000 common word pairs, which restricts its coverage and, consequently, limits the performance of our model. This limitation has been addressed by ExtViSentiWordNet, which provides enhanced coverage for the Vietnamese lexicon.

## 6. Conclusion

In this paper, we present a method for extending Vietnamese SentiWordNet to enhance its coverage of the Vietnamese lexicon. Additionally, we propose an integrated approach for Vietnamese sentiment analysis that combines both distributional and lexical semantic features. Specifically, we introduce CombViSA, a sentiment analysis model that leverages the PhoBERT-V2 pre-trained model, which provides robust optimization for the Vietnamese BERT model, along with lexical features derived from the expanded Vietnamese SentiWordNet. Experimental results on the VLSP 2016 and AIVIVN 2019 datasets demonstrate a significant improvement in performance with our model. In the future, we plan to apply ExtViSentiWordNet to various NLP tasks, including product brand analysis and sentence similarity.

## References

[1] D. Q. Nguyen and A. Tuan Nguyen, "PhoBERT: Pre-trained language models for Vietnamese", *in Findings of the Association for Computational Linguistics: EMNLP 2020, T. Cohn, Y. He, and Y. Liu, Eds. Online: Association for Computational Linguistics*, Nov. 2020, pp. 1037–1042.

[2] A. Esuli and F. Sebastiani, "SENTIWORDNET: A publicly available lexical resource for opinion mining", *in Proceedings of the Fifth International Conference on Language Resources and Evaluation (LREC'06). Genoa, Italy: European Language Resources Association (ELRA)*, May 2006.

[3] R. Feldman, "Techniques and applications for sentiment analysis", Communications of the ACM, vol. 56, no. 4, pp. 82–89, apr 2013.

[4] M. Ghiassi, J. Skinner, and D. Zimbra, "Twitter brand sentiment analysis: A hybrid system using n-gram analysis and dynamic artificial neural network", *Expert Syst. Appl.*, vol. 40, no. 16, pp. 6266–6282, 2013.

[5] A. Abbasi and H. Chen, "Applying authorship analysis to extremistgroup web forum messages", *IEEE Intelligent Systems*, vol. 20, no. 5, pp. 67–75, 2005.

[6] E. Macovei, "Neo-nazis sympathizers on the forums of the romanian online publications", *Styles of Communication*, vol. 5, no. 1, pp. 93–109, 2013.

[7] J. F. K. L. M. J. Cohen, K., "Detecting linguistic markers for radical violence in social media," *Terrorism and Political Violence*, vol. 26, pp. 246–256, 2014.

[8] B. Pang, L. Lee, and S. Vaithyanathan, "Thumbs up? sentiment classification using machine learning techniques", *in Proceedings of the 2002 Conference on Empirical Methods in Natural Language Processing (EMNLP 2002)*, Jul. 2002, pp. 79–86.

[9] B. M. and V. B., "Sentiment analysis using support vector machine based on feature selection and semantic analysis", *International Journal of Computer Applications*, vol. 146, pp. 26–30, 07 2016.

[10] P. P. Surya and B. Subbulakshmi, "Sentimental analysis using naive bayes classifier", *in 2019 International Conference on Vision Towards Emerging Trends in Communication and Networking (ViTECoN)*, 2019, pp. 1–5.

[11] M. Taboada, J. Brooke, M. Tofiloski, K. Voll, and M. Stede, "Lexiconbased methods for sentiment analysis", *Computational Linguistics*, vol. 37, pp. 267–307, 06 2011.

[12] Q. Le and T. Mikolov, "Distributed representations of sentences and documents", *International Conference on Machine Learning*, ICML 2014, vol. 4, 05 2014.

[13] X. Zhang, J. Zhao, and Y. Lecun, "Character-level convolutional networks for text classification", *Proceedings of the 29th International Conference on Neural Information Processing Systems,* Vol. 1, 2015.
[14] O. Araque, I. Corcuera-Platas, J. F. Sánchez-Rada, and C. A. Iglesias, "Enhancing deep learning sentiment analysis with ensemble techniques in social applications", *Expert Systems with Applications*, vol. 77, pp. 236–246, 2017.
[15] M. Amjad, A. F. Gelbukh, I. Voronkov, and A. Saenko, "Comparison of text classification methods using deep learning neural networks", *Lecture Notes in Computer Science*, vol. 13452. Springer, 2019, pp. 438–450.
[16] P. Tiwari, H. M. Pandey, A. Khamparia, and S. Kumar, "Twitter-based opinion mining for flight service utilizing machine learning", *Informatica (Slovenia)*, vol. 43, no. 3, 2019.
[17] Y. Zhang, Z. Zhang, D. Miao, and J. Wang, "Three-way enhanced convolutional neural networks for sentence-level sentiment classification", *Inf. Sci.*, vol. 477, pp. 55–64, 2019.
[18] A. Yadav and D. K. Vishwakarma, "Sentiment analysis using deep learning architectures: a review", *Artif. Intell. Rev.*, vol. 53, no. 6, pp. 4335–4385, 2020.
[19] O. Habimana, Y. Li, R. Li, X. Gu, and G. X. Yu, "Sentiment analysis using deep learning approaches: an overview", *Science China Information Sciences*, vol. 63, 2019.
[20] W. Etaiwi, D. Suleiman, and A. Awajan, "Deep learning based techniques for sentiment analysis: A survey", *Informatica*, vol. 45, pp. 89–96, 08 2021.
[21] J. Li, Y. Wang, and J. Wang, "An analysis of emotional tendency under the network public opinion: Deep learning", *Informatica (Slovenia)*, vol. 45, no. 1, 2021.
[22] A. Al-Rasheed, "Finding influential users in social networking using sentiment analysis", *Informatica*, vol. 46, 03 2022.
[23] J. Devlin, M. Chang, K. Lee, and K. Toutanova, "BERT: pre-training of deep bidirectional transformers for language understanding", *CoRR*, vol. abs/1810.04805, 2018.
[24] Y. Liu, M. Ott, N. Goyal, J. Du, M. Joshi, D. Chen, O. Levy, M. Lewis, L. Zettlemoyer, and V. Stoyanov, "Roberta: A robustly optimized bert pretraining approach", *ArXiv*, vol. abs/1907.11692, 2019.
[25] S. Hochreiter and J. Schmidhuber, "Long short-term memory", *Neural computation*, vol. 9, no. 8, pp. 1735–1780, 1997.
[26] S. Lai, L. Xu, K. Liu, and J. Zhao, "Recurrent convolutional neural networks for text classification", *in AAAI*, vol. 333, 2015, pp. 2267–2273.
[27] G. A. Miller, R. Beckwith, C. Fellbaum, D. Gross, and K. Miller, "Wordnet: An on-line lexical database", *International Journal of Lexicography*, vol. 3, pp. 235–244, 1990.
[28] P. D. Turney and M. L. Littman, "Measuring praise and criticism: Inference of semantic orientation from association" *CoRR*, vol. cs.CL/0309034, 2003.
[29] X.-S. Vu, H.-J. Song, and S.-B. Park, "Building a vietnamese sentiwordnet using vietnamese electronic dictionary and string kernel", *Lecture Notes in Computer Science*, vol. 8863. Springer, 2014, pp. 223–235.
[30] Q. T. Nguyen, T. L. Nguyen, N. H. Luong, and Q. H. Ngo, "Finetuning bert for sentiment analysis of vietnamese reviews", *CoRR*, vol. abs/2011.10426, 2020.
[31] N. T. Q. V. X. L. T. M. V. N. X. B. L. A. C. NGUYEN THI MINH HUYEN, NGUYEN VIET HUNG, "Vlsp shared task sentiment analysis", *Journal of Computer Science and Cybernetics*, vol. V.34, N.4, 2018.